# Fuzzy Clustering Data Given in the Ordinal Scale

**Zhengbing Hu**
School of Educational Information Technology, Central China Normal University, Wuhan, China
Email: hzb@mail.ccnu.edu.cn

**Yevgeniy V. Bodyanskiy**
Kharkiv National University of Radio Electronics, Kharkiv, Ukraine,
Email: yevgeniy.bodyanskiy@nure.ua

**Oleksii K. Tyshchenko and Viktoriia O. Samitova**
Kharkiv National University of Radio Electronics, Kharkiv, Ukraine,
Email: lehatish@gmail.com, samitova@ gmail.com

*Abstract* — A fuzzy clustering algorithm for multidimensional data is proposed in this article. The data is described by vectors whose components are linguistic variables defined in an ordinal scale. The obtained results confirm the efficiency of the proposed approach.

## I. Introduction

That's a very common situation for sociological, medical and educational tasks when initial data is given not in a numeric scale but rather in a rank (ordinal) scale [1-5]. This information (for the one-dimensional case) is given in the form of an ordered sequence of linguistic variables $x^1, x^2, ..., x^i, ..., x^m$, $1 < ... < j-1 < j < j+1 < ... < m$ where $x^j$ is a linguistic variable and $j$ is its corresponding rank.

A typical example is a traditional (for Ukraine and some other ex-USSR countries) educational grading system like "poor", "fair", "good", "excellent". Let's note that a person in its daily activities is much more likely to use an ordinal scale rather than a numerical one.

The simplest approach for solving these clustering problems on an ordinal scale is based on replacement of linguistic variables with their ranks but this method is incorrect in most cases because it assumes equal distances between neighboring numerical ranks [6-12]. It is intuitively clear that a distance between "poor" and "fair" is much longer than a distance between "fair" and "good" while assessing students' knowledge. Many similar examples can be found in medicine.

That's a more natural approach that is based on fuzzification of input data and further usage of fuzzy clustering methods. Thus, an initial set of linguistic variables $x^1, x^2, ..., x^i, ..., x^m$ is replaced with a set of membership functions $\mu_1(x), \mu_2(x), ..., \mu_m(x)$ defined in the interval $[0, 1]$. This method was used in [13] where clustering (based on the Fuzzy C-means (FCM) algorithm [14]) was not performed for the initial data but for parameters describing corresponding membership functions, although a method for determining these parameters was not specified.

There's an approach that looks more natural. It's developed by R.K. Brouwer [15-21] and based on the frequency distribution analysis of the specific values' occurrence of linguistic variables. A limitation of this approach is an assumption about the Gaussian distribution of the initial data. This assumption is not usually met in many real-world applications.

The initial data for solving the task is a sample of observations which contains $N$ $n-$dimensional feature vectors $X = \{x(1), x(2), ..., x(k), ..., x(N)\}$, $k = 1, 2, ..., N$, $x(k) = \{x_i^j(k)\}$,

$i = 1, 2, ..., n$; $j = 1, 2, ..., m$ is a rank of a specific value of a linguistic variable in the $i$ – th coordinate of the $n$ – dimensional space for the $k$ – th object to be clustered.

A result of this algorithm is partition of the initial dataset $X$ into $m$ classes (clusters) as well as calculation of a membership level $w_j(k)$ $k$ – th feature vector to the $j$ – th cluster.

The remainder of this paper is organized as follows: Section 2 describes a procedure of the initial data fuzzification. Section 3 describes. a fuzzy clustering method for ordinal data. Section 4 describes experimental results. Conclusions and future work are given in the final section.

## II. THE INITIAL DATA FUZZIFICATION

A fuzzification process for a sequence of rank linguistic variables may be considered by an example of the one-dimensional sample $x(1), x(2), ..., x(N)$ where each observation $x(k)$ may be ascribed one of the $j$ ranks, $j = 1, 2, ..., m$.

Let the value $x(k)$ (which corresponds to the $j$ – th rank) be seen in the sample $N_j$ times. Then a relative frequency of occurrence of the $j$ – th rank can be introduced

$$f_j = \frac{N_j}{N}$$

as well as cumulative frequencies

$$F_1 = \frac{f_1}{2}, \quad F_j = \frac{f_j}{2} + \sum_{l=1}^{j-1} f_l, \quad j = 2, 3, ..., m$$

while naturally there's a condition

$$\sum_{i=1}^{m} f_i = 1.$$

Based on these cumulative frequencies, membership functions' $\mu_j(x)$ centers are formed (Fig.1). At the same time, it's convenient to use a recurrence relation to compute centers

$$c_1 = 0,5 f_1, \quad c_j = c_{j-1} + 0,5(f_{j-1} + f_j), \quad j = 2, 3, ... m,$$

and to set membership functions in the form

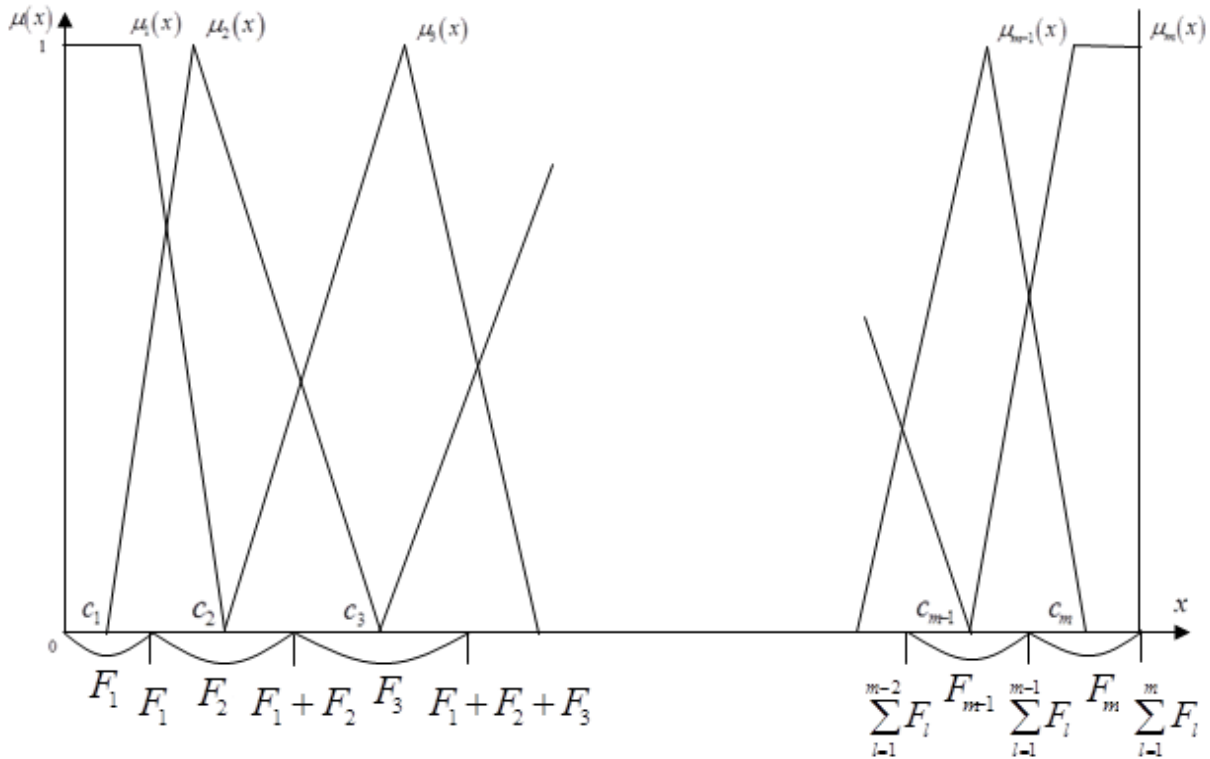

Fig.1. Membership functions for a set of rank variables

$$\mu_1(x) = 1,\ x \in [0, c_1],$$

$$\mu_j(x) = \begin{cases} \dfrac{x - c_{j-1}}{c_j - c_{j-1}}, & x \in [c_{j-1}, c_j], \\ \dfrac{c_{j+1} - x}{c_{j+1} - c_j}, & x \in [c_j, c_{j+1}], \\ 0,\ x \notin [c_{j-1}, c_{j+1}], \end{cases}$$

$$\mu_m(x) = 1,\ x \in [c_m, 1].$$

This method of setting membership functions automatically provides the unity (Ruspini) partition which means that the following condition is fulfilled

$$\sum_{j=1}^{m} \mu_i(x) = 1.$$

Although it's possible to use functions of some other view with a finite support

$$\mathrm{supp}\,\mu_j(x) = [c_{j-1}, c_{j+1}].$$

Let's consider two neighboring membership functions $\mu_j(x)$ and $\mu_{j+1}(x)$ (Fig.2). Using a term of the $\alpha -$ cut

$$A_\alpha = \{x \in X : \mu(x) \geq \alpha\},$$

an area of influence for two neighboring ranks (shaded in Fig.2) can be introduced in the form of

$$\begin{cases} A_j^R = \left\{ x \in [c_j, c_j + 0,5 f_j] : \mu_j(x) \geq \alpha_j^R \right\} = \\ \quad = 1 - 0,5 \dfrac{f_j}{c_{j+1} - c_j}, \\ A_{j+1}^L = \left\{ x \in [c_{j+1} - 0,5 f_{j+1}, c_{j+1}] : \mu_{j+1}(x) \geq \alpha_{j+1}^L \right\} = \\ \quad = 1 - 0,5 \dfrac{f_{j+1}}{c_{j+1} - c_j} \end{cases} \quad (1)$$

where $L$ and $R$ stand for the left and right sides of neighboring membership functions. If there's a hit of some observation in the area of influence of a particular rank, we can talk about crisp belonging to this rank.

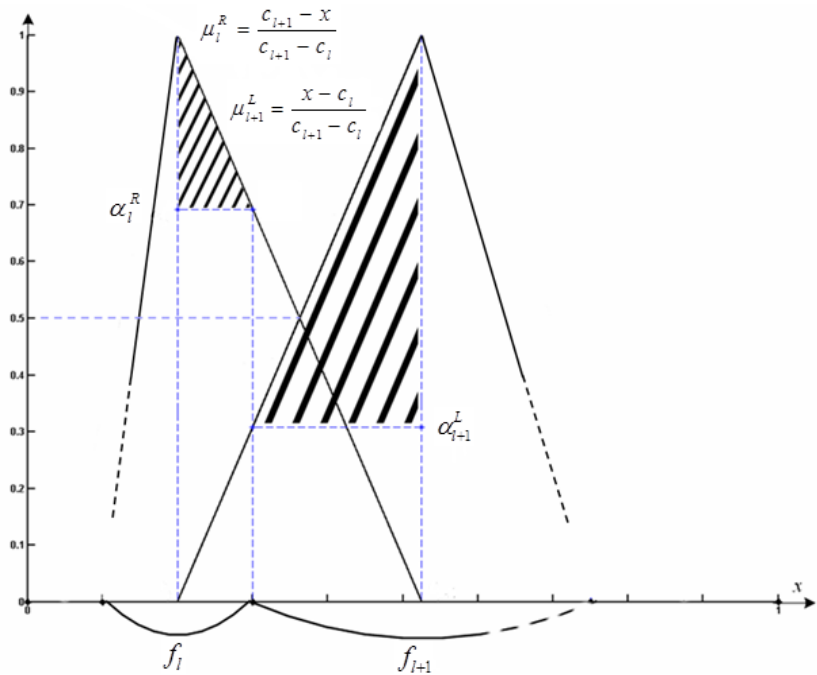

Fig.2. The areas of influence for neighboring ranks.

## III. A FUZZY CLUSTERING METHOD FOR ORDINAL DATA

Since we have a multidimensional data sample of observations (vectors) to be clustered, the fuzzification procedure should be performed in every coordinate of the $n-$dimensional feature space similarly to the previous example. $nm$ membership functions are formed (they have centers $c_{ij}$) during this procedure like it's shown for a two-dimensional case (Fig.3). Here comes an object $x(k)$ to be clustered with coordinates

$$x(k) = \begin{pmatrix} x_1^2(k) \equiv a \\ x_2^4(k) \equiv e \end{pmatrix}.$$

As one can see in Fig.3, membership functions $\mu_{12}(x_1), \mu_{13}(x_1), \mu_{14}(x_1), \mu_{22}(x_2), \mu_{23}(x_2), \mu_{24}(x_2)$ work in this case in such a way that it's rather hard to make an immediate decision about belonging of $x(k)$ to one of the classes «Poor», «Fair», «Good» and «Excellent» (Fig.3). A process of fuzzy clustering for rank variables is carried out at the same example that is shown in Fig.4. After $nm$ membership functions (in our example, $2 \times 4 = 8$ membership functions) have been formed, $n-$dimensional vectors (cluster centroids) $c_j = (c_{1j}, c_{2j}, ..., c_{nj})^T$, $j = 1, 2, ..., m$ are taken into consideration (in our example, $c_1 = (c_{11}, c_{12})^T, c_2 = (c_{21}, c_{22})^T, c_3 = (c_{31}, c_{32})^T, c_4 = (c_{41}, c_{42})^T$) with their areas of influence which are described by ratios (1) (shaded areas in the picture). If an object gets into these regions, we can say that there's crisp membership of an object $x(k)$ to a specific cluster. We have an object $x(k) = (e, a)^T$ to be classified which is represented in a numerical form with coordinates $c_{12}$ and $c_{24}$ after fuzzification..

Then distances $d(x(k), c_i) = \|x(k) - c_i\|$ between $x(k)$ and all centroids $c_i$ are calculated. Membership levels $w_j(k)$ of a vector $x(k)$ to the $j-$th cluster should be defined according to the FCM procedure [22]

$$w_j(k) = \frac{\|x(k) - c_j\|^{-2}}{\sum_{l=1}^{m} \|x(k) - c_l\|^{-2}} = \frac{d^{-2}(x(k), c_j)}{\sum_{l=1}^{m} d^{-2}(x(k), c_l)}. \qquad (2)$$

A drawback of the estimate (2) is the fact that an object (except the case when $x(k)$ gets into an area of influence of a centroid) equally belongs to all the existing clusters which leads to loss of a physical sense in the rank scale. So, an object under consideration $x(k) = (e, a)^T$ with a non-zero membership level may belong both to the «Excellent» cluster and to the «Poor» cluster. Obviously, that doesn't make any sense.

It seems reasonable in this regard to rank all the distances $d(x(k), c_j)$ (which have been previously computed) in an ascending order and to choose the minimum distance $d_{\min\min}(x(k), c_j)$ and the one following it $d_{\min}(x(k), c_l)$. Then we can use the expression (2) with the only difference that we take into account only the two minimum distances. Finally, $x(k)$ belongs to two neighboring clusters with centroids $c_j$ and $c_{j+1}$ (or $c_{j-1}$) with some membership levels $w_j(k)$ and $w_{j+1}(k)$ (or $w_{j-1}(k)$).

Thus, the fuzzy clustering algorithm for multidimensional observations given in the ordinal scale is implemented as a sequence of steps:

1. Calculation of relative $f_j$ and cumulative $F_j$ frequencies in a dataset $x(1), x(2), ..., x(k), ..., x(N)$;

2. Fuzzification of an initial data set of linguistic variables by forming $mn$ membership functions $\mu_{ij}(x_i)$, $j = 1, 2, ..., m; i = 1, 2, ..., n$ and $m$ vectors–centroids $c_j = (c_{1j}, c_{2j}, ..., c_{nj})^T$ for the clusters to be formed;

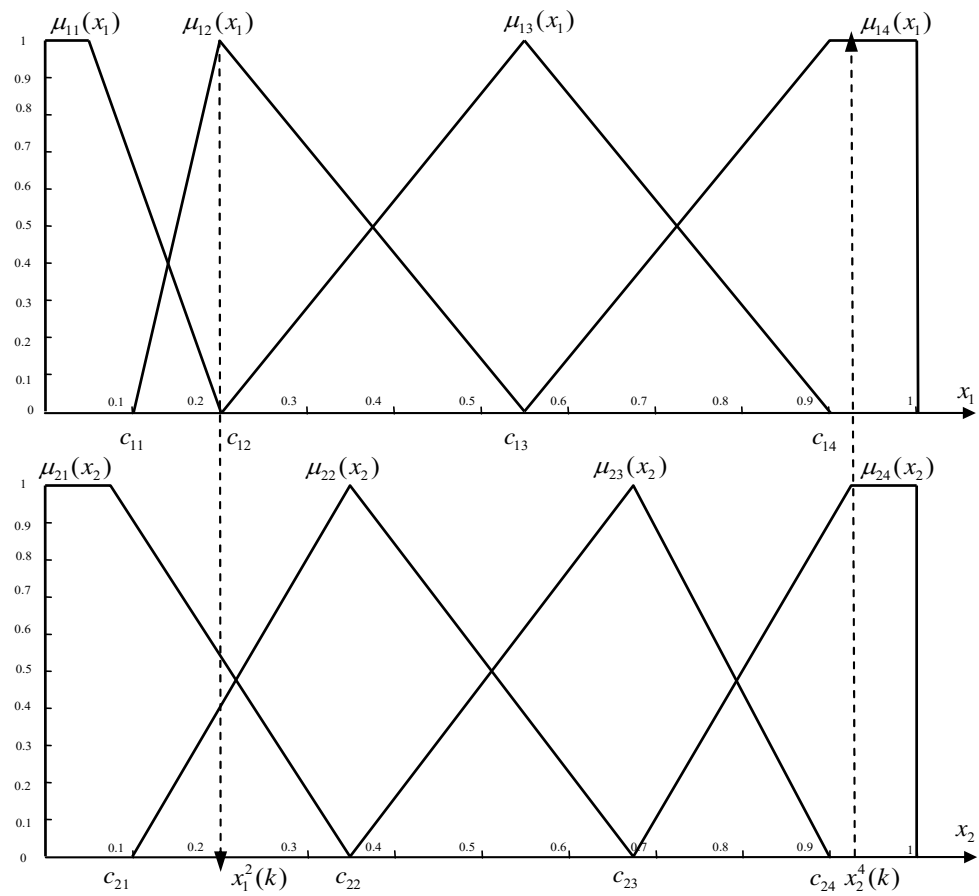

Fig.3. Membership functions of a 2D feature space

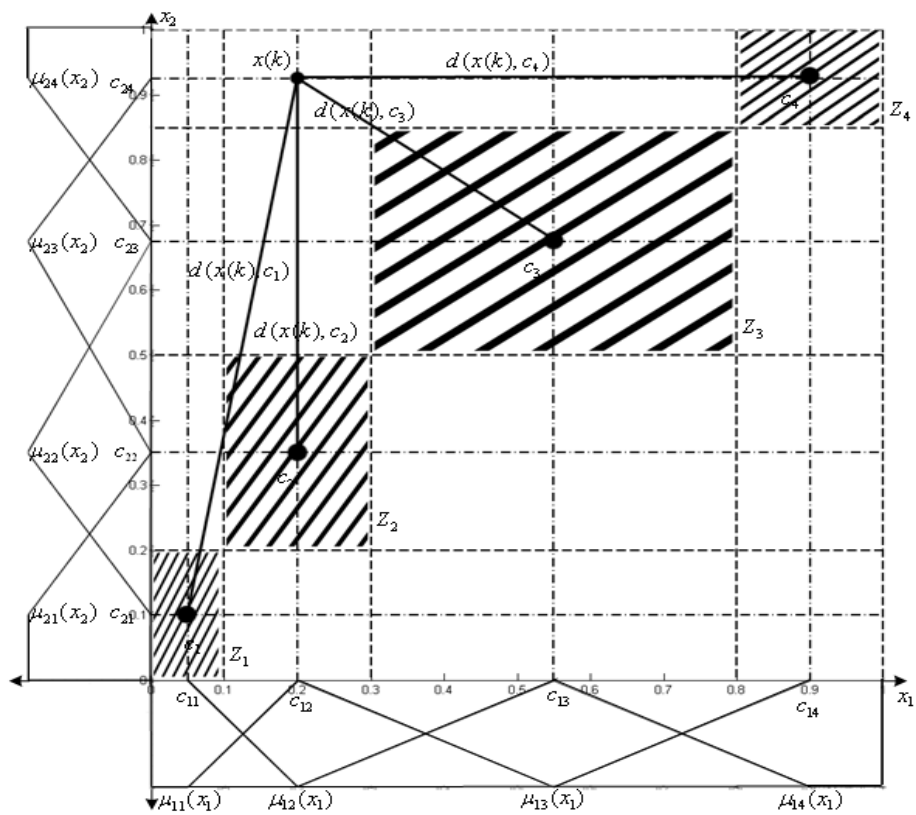

Fig.4. Fuzzy clustering for rank variables

3. Building areas of influence $Z_j$ for centroids $c_j$ in the form of an orthotop with edges $c_{ij} \pm 0{,}5 f_{ij}$

4. Checking a possibility of crisp clustering in the form: if $x(k) \in Z_j$, this observation can be unambiguously classified, i.e. $w_j(k) = 1$ and $w_j(k) = 0$ for all other $l \neq k$;

5. If the previous condition is not fulfilled, calculation of all distances $d(x(k), c_j) = \|x(k) - c_j\|$ is performed;

6. Choosing two minimal distances $d_{\min\min}(x(k), c_j)$ and $d_{\min}(x(k), c_l)$ where $l$ may take on a value of $j-1$ or $j+1$;

7. Computing membership levels for $x(k)$ to find out its belonging to two neighboring clusters:

$$w_j(k) = \frac{d^{-2}_{\min\min}(x(k), c_j)}{d^{-2}_{\min\min}(x(k), c_j) + d^{-2}_{\min}(x(k), c_l)},$$

$$w_l(k) = \frac{d^{-2}_{\min}(x(k), c_l)}{d^{-2}_{\min\min}(x(k), c_j) + d^{-2}_{\min}(x(k), c_l)}.$$

IV. EXPERIMENTS

To check the efficiency of the proposed method, we have gathered data about students' academic performance at one faculty at Kharkiv National University of Radio Electronics. A dataset contains students' grades in 6 subjects for 135 persons.

Statistical analysis showed that a hypothesis for each variable (subject) about the estimates' normal distribution was not confirmed (Fig.5).

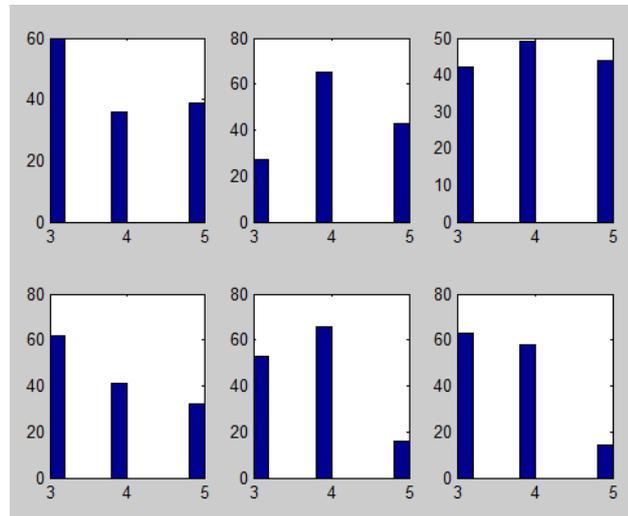

Fig.5. A histogram of grades' distribution for six subjects

Having applied the proposed method, we calculated centroids for each rank (grades) for each variable and built membership functions with areas of influence (shown in Fig.6-11).

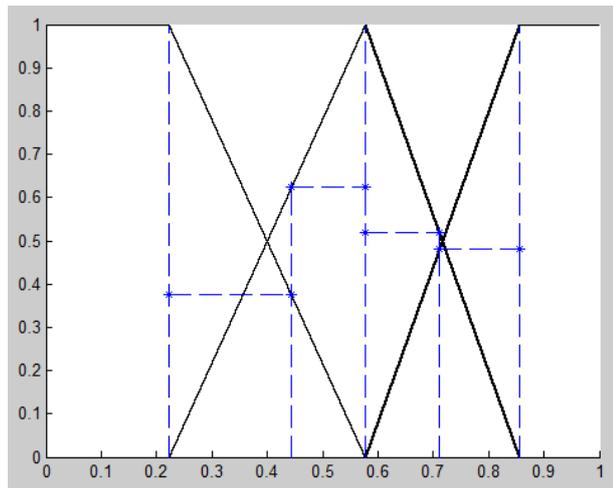

Fig.6. Membership functions with $\alpha$ – cuts for 6 subjects (case 1)

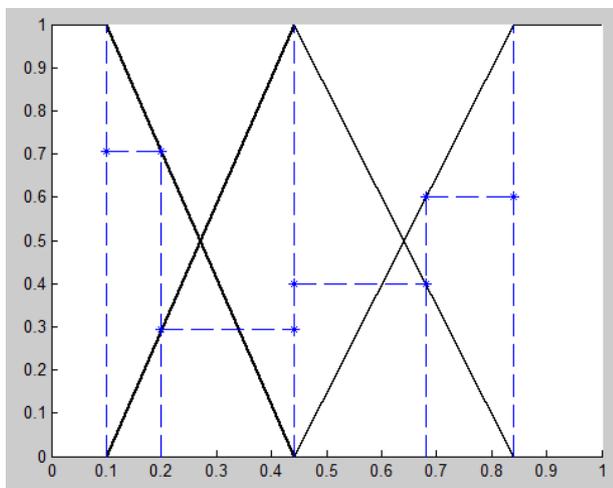

Fig.7. Membership functions with $\alpha$ – cuts for 6 subjects (case 2)

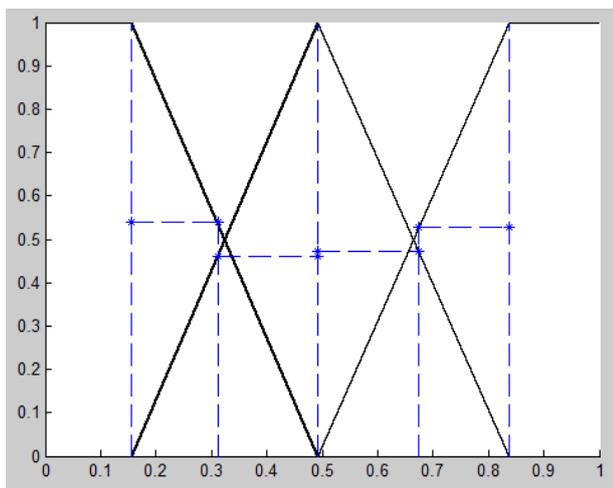

Fig.8. Membership functions with $\alpha$ – cuts for 6 subjects (case 3)

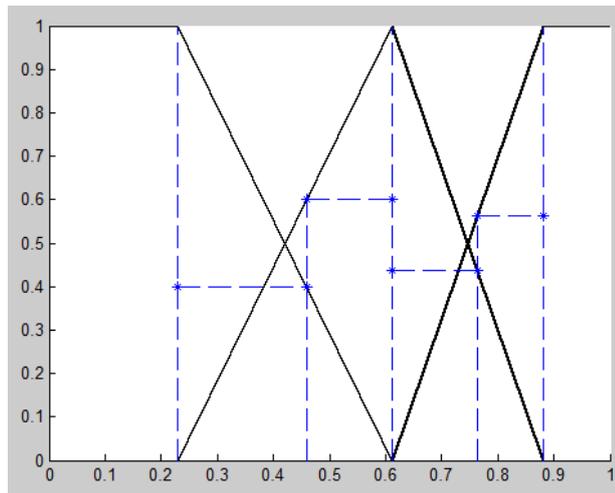

Fig.9. Membership functions with $\alpha$ – cuts for 6 subjects (case 4)

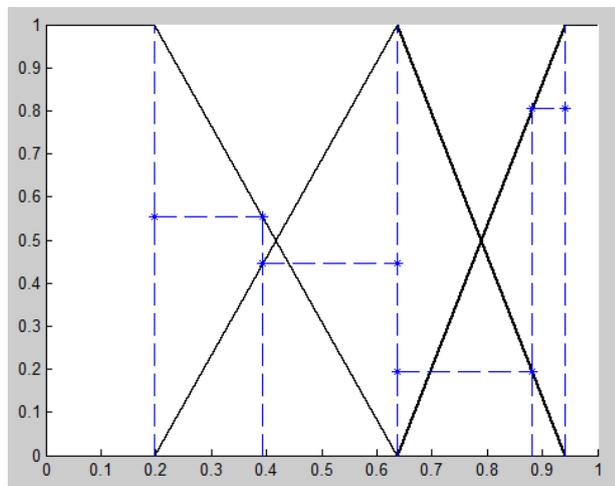

Fig.10. Membership functions with $\alpha$ – cuts for 6 subjects (case 5)

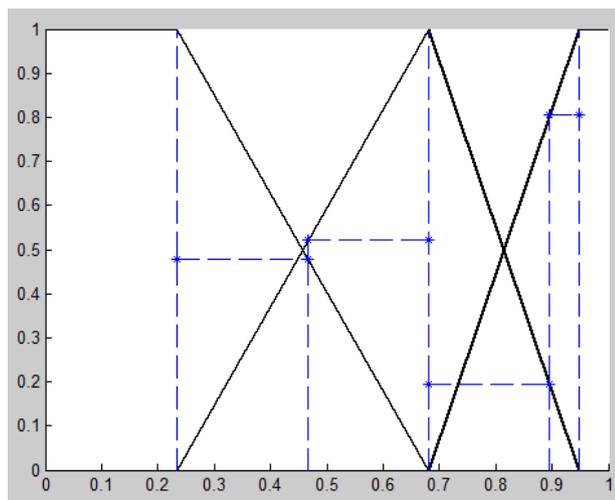

Fig.11. Membership functions with $\alpha$ – cuts for 6 subjects (case 6)

The proposed algorithm was compared to the FCM procedure $(\beta = 2)$ and the clustering method for ordinal data by R.K. Brouwer (BFCM) [15]. Since classes weren't initially given in the dataset, it's quite difficult to talk about a clustering accuracy of each method. Although analyzing the computed membership functions for some observations (Tables 1 and 2), it should be noticed that the proposed approach performs data processing in a more correct manner. It's clear that the demonstrated observations (Tables 1 and 2) may belong to the "Good" and "Excellent" clusters with a certain membership level but there's no way for them to belong to the "Fair" cluster. It's proven by the FCM and BFCM methods (Table 3).

Table 1. Observations from the dataset. Examples (Subjects #1 - #3)

| # obs. | Subject #1 | Subject #2 | Subject #3 |
|---|---|---|---|
| 1 | Fair | Excellent | Excellent |
| 2 | Fair | Excellent | Fair |

Table 2. Observations from the dataset. Examples (Subjects #4 - #6)

| # obs. | Subject #4 | Subject #5 | Subject #6 |
|---|---|---|---|
| 1 | Excellent | Excellent | Excellent |
| 2 | Good | Fair | Fair |

Table 3. Membership levels of observations' belonging to clusters

| Models | # obj. | Fair | Good | Excellent |
|---|---|---|---|---|
| FCM | 1 | 0.13 | 0.37 | 0.5 |
|  | 2 | 0.35 | 0.48 | 0.17 |
| BFCM | 1 | 0.14 | 0.26 | 0.6 |
|  | 2 | 0.31 | 0.49 | 0.2 |
| MBFCM | 1 | 0 | 0.38 | 0.62 |
|  | 2 | 0.56 | 0.44 | 0 |

V. CONCLUSION

The task of fuzzy clustering ordinal data has been solved with the help of the proposed method. The most peculiar feature of this procedure is the fact that it can work under conditions when objects belong to different clusters at the same time.

ACKNOWLEDGMENT

This scientific work was supported by RAMECS and CCNU16A02015.